# Systematic Literature Review of Vision-Based Approaches to Outdoor Livestock Monitoring with Lessons from Wildlife Studies

Review of Vision-Based Approaches to Outdoor Livestock Monitoring

STACEY D. SCOTT*

School of Computer Science, University of Guelph, stacey.scott@uoguelph.ca

ZAYN J. ABBAS

School of Computer Science, University of Guelph

FEERASS ELLID

School of Computer Science, University of Guelph

ELI-HENRY DYKHNE

School of Computer Science, University of Guelph

MUHAMMAD MUHAIMINUL ISLAM

School of Computer Science, University of Guelph

WEAM AYAD

School of Computer Science, University of Guelph

KRISTINA KACMOROVA

School of Computer Science, University of Guelph

DAN TULPAN

Department of Animal Biosciences, University of Guelph

MINGLUN GONG

School of Computer Science, University of Guelph

Precision livestock farming (PLF) aims to improve the health and welfare of livestock animals and farming outcomes through the use of advanced technologies. Computer vision, combined with recent advances in machine learning and deep learning artificial intelligence approaches, offers a possible solution to the PLF ideal of 24/7 livestock monitoring that helps facilitate early detection of animal health and welfare issues. However, a significant number of livestock species are raised in large outdoor habitats that pose technological challenges for computer vision approaches. This review provides a comprehensive overview of computer vision methods and open challenges in outdoor animal monitoring. We include research from both the livestock and wildlife fields in the review because of the similarities in appearance, behaviour, and habitat for many livestock and wildlife. We focus on large terrestrial mammals, such as cattle, horses, deer, goats, sheep, koalas, giraffes, and elephants. We use an image processing pipeline to frame our discussion and highlight the current capabilities and open technical challenges at each stage of the pipeline. The review found a clear trend towards the use of deep learning approaches for animal detection, counting, and multi-species classification. We discuss in detail the applicability of current vision-based methods to PLF contexts and promising directions for future research.

---

* University of Guelph, 50 Stone Road East, Guelph Ontario, Canada N1G 2W1

## 1 INTRODUCTION

Precision livestock farming (PLF) concerns the use of advanced technologies to improve the productivity, health and welfare of individual livestock animals and to optimize farming practices [6,10,60]. Early detection and treatment of health and welfare issues is a common goal of PLF applications [11]. However, this is highly challenging to apply in outdoor pastoral or open range environments where livestock are often out of view and out of range of the power or telecommunications infrastructure upon which many current PLF animal monitoring systems rely. Thus, animal monitoring systems such as wearable activity collars are often not appropriate in these contexts.

Instead, PLF researchers are beginning to explore remote sensing options, such as animal surveys using uninhabited aerial vehicles (UAVs), or drones. Commercial UAVs currently exist marketed to livestock farmers (e.g., cattle ranchers) to collect visual data on their farmland and animals. For example, commercially available UAVs can be used to find lost cattle or detect damaged fences [33]. However, current UAV solutions require farmers to remotely fly and/or manually view video feeds to determine livestock condition or location. Unlike other PLF technologies that use automated sensors to detect animal health and welfare issues, this approach is unreliable and unlikely to detect issues in a timely manner. Ideally, UAVs or other vision-based sensors would be able to survey an area autonomously, automatically detect health or welfare issues, and inform the farmer if any issues were detected.

Given recent advances in computer vision-based solutions in other domains, especially based on machine and deep learning technologies [19], we were curious about how close research solutions were to supporting this ideal PLF scenario, whether it might be feasible in the near future, and what potential barriers exist to a fully automated vision-based remote sensing animal monitoring solution. To explore these questions, we conducted a systematic literature review on vision-based approaches to animal monitoring in outdoor pastoral or open range habitats. Given the similarities between many livestock and wildlife in appearance, behaviour, and also habitat, we expanded our survey to include literature in both the livestock and wildlife research fields. While the goals of wildlife researchers may differ from those of livestock researchers, we felt many of the technical issues of animal monitoring in both contexts were sufficiently similar to help us gain useful insights into the opportunities and limitations of vision-based animal monitoring from both bodies of literature.

This literature review provides an introduction to the current autonomous computer vision methods being used in the livestock and wildlife fields and serves as a guide to the computer vision process in precision livestock farming. We propose an image processing pipeline, and then break down each pipeline stage to help explain the current state-of-the-art methods and identify suggestions for future improvements. The key contributions of the paper include:

- An overview of current livestock and wildlife monitoring trends related to research methods, data types, animal monitoring tasks, and contributing countries.
- A systematic categorization of computer vision methods used in outdoor livestock and wildlife monitoring at each stage of a typical image processing pipeline.
- A detailed discussion of the technical capabilities and challenges of computer vision methods used in outdoor livestock and wildlife monitoring.
- A reflection on the feasibility of current computer vision methods for the PLF domain.
- A set of recommendations for future research to improve vision-based outdoor livestock monitoring.

## 2 RESEARCH QUESTIONS AND IMAGE PROCESSING PIPELINE

To understand the feasibility of using automated remote sensing techniques to facilitate outdoor livestock monitoring, we conducted a systematic literature review focused on the following research questions:

- What computer vision based techniques have been used to monitor livestock or wildlife in outdoor contexts?
- What are the open challenges to creating feasible vision-based automated livestock monitoring solutions?

To explore these questions, we examined the literature through the lens of an image processing pipeline (Figure 1), which defines the workflow steps involved in many computer vision-based approaches to process image data during automated image analysis solutions. We use this pipeline as a framework to help understand the unique challenges to applying vision-based approaches in animal monitoring contexts, and what advances livestock and wildlife researchers have made thus far in addressing these challenges.

As Figure 1 shows, the image processing pipeline often involves five stages: gathering input data, pre-processing, selecting areas of interest, processing of selected areas, and decision making [36,55,92]. We use this image processing pipeline as a framework to highlight the research contributions towards computer vision approaches for outdoor livestock monitoring. Not every surveyed approach utilizes all five stages. We often found blurred lines between adjacent stages. Vision-based approaches that use machine learning (ML) or deep learning (DL) often internalize the functionality of multiple pipeline stages into a single closed-box step. We aim to highlight the progress and challenges in using vision-based approaches within PLF by discussing the literature in the context of this pipeline.

To provide context for our literature analysis, we describe each pipeline stage below. Note, this pipeline is not always a linear process, and stages can be repeated or be done out of order to improve results in a given context.

The **Input Data** stage involves gathering the image data to be processed and/or to be used for training and testing ML or DL algorithms. Data can be captured in the form of still or video imagery in the desired color spectra using stationary cameras, cameras attached to aircraft, reused from existing datasets, and even generated artificially (e.g., data augmentation). These methods and their applications are discussed in section 6.1.

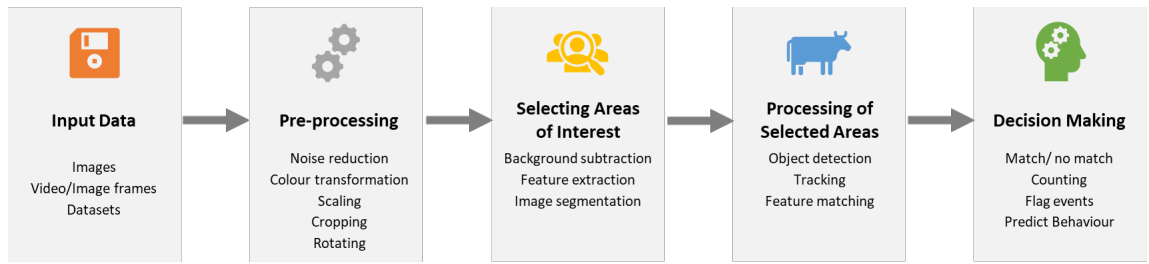

Figure 1: Image Processing Pipeline: stages involved in processing imagery data.

Identifying regions of importance within an image is done in the **Selecting Areas of Interest** stage. This helps reduce the computational search space and complexity for algorithms applied in later stages. Techniques discussed in section 6.3 help distinguish between features of interest (e.g., animal) and the background context of a digital image.

During the **Processing of Selected Areas** stage, further analysis of the selected areas of interest is performed to detect and label target objects (e.g., cow, sheep). Generally, the areas of interest are classified by user-defined categories. Techniques used within this stage include traditional computer vision algorithms, ML, and DL techniques, as discussed in section 6.4.

The **Decision-Making** stage is the final stage of the image processing pipeline. It involves the interpretation, inference, and use of classified images. Categories within this stage are based on research goals as well as PLF applications. These include detection, multi-species identification, counting, and tracking, as discussed in section 6.5.

## 3 RESEARCH METHODOLOGY

We followed the PRISMA guidelines and methodology [74] for transparent reporting of systematic literatures reviews. The Web of Science (WoS) and EBSCO Host source databases were used. The initial data search was done in June 2020 and a second search to include newer papers was done in June 2021, with forward and backward snowballing occurring until article submission. The search terms are described below along with our inclusion and exclusion criteria for building our data corpus. The Mendeley[1] online reference manager tool was used to facilitate the literature review process and to share the data corpus among the research team.

---

[1] http://www.mendeley.com

## 4 SEARCH TERMS

Literature review search terms were generated and iterated on to produce as many relevant articles as possible. The specific search terms used in both the Web of Science (WoS) and EBSCO Host databases are provided below:

*"(Comput\* OR Machine OR Algorithm OR Neural OR Processing OR artificial) AND (Aerial OR Outside OR Outdoors OR Remote Sens\*) AND (Sensor\* OR Detect\* OR Monitor OR Camera\* OR Drone OR UAV OR UAS) AND (Vision OR Imag\*) AND (Livestock OR Bovine OR Pig OR Cow OR Herd OR Animal OR Cattle OR wildlife ) AND (farm\* or agri\* or ecolog\* or wildlife) NOT (Crop\* OR Plant\* OR Bird OR Rodent OR "human health" OR Microb\* OR Marine OR Insect)"*

## 5 INCLUSION AND EXCLUSION CRITERIA

The articles were filtered to include peer-reviewed journal or conference papers published in 2005 or later. This timeframe was chosen since initial research showed only a few papers prior to 2005. Earlier papers were captured in our snowballing efforts. Papers were then filtered manually to remove papers with the following exclusion criteria:

- Extended abstracts.
- Papers lacking an English translation.

Papers were also filtered according to the following inclusion criteria:

- Use of computer vision technologies, or those that directly support computer vision.
- Topic must involve livestock or wildlife in an outdoor, land-based setting.

The initial search found 171 papers in WoS and 244 papers in EBSCO. After removing duplicates, papers were filtered using the inclusion and exclusion criteria based on review of their abstracts, resulting in 96 papers. The screening process was repeated based on a review of the full manuscript, resulting in 38 papers. The "snowball" method was then applied to source additional key papers: Papers cited within the original 38 papers that appeared both relevant and passed the inclusion criteria, were added to our data corpus, adding an additional 12 papers. This selection process is summarized in the PRISMA flow diagram shown in Figure 2.

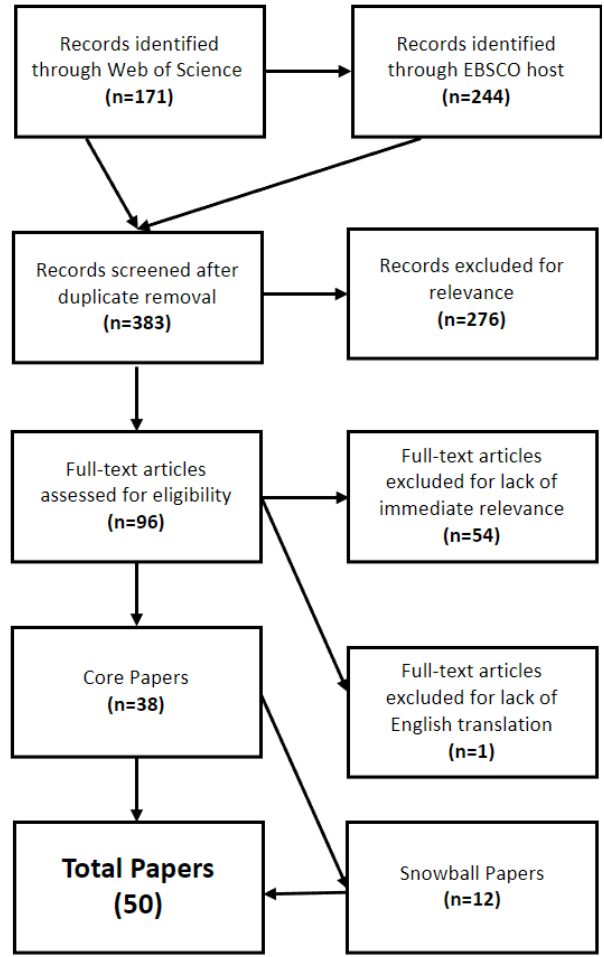

Figure 2. PRISMA flow diagram.

## 6 DATA CORPUS

The data corpus was split into two main categories: prior literature reviews and studies. A summary table of the papers in our data corpus is included in the Digital Appendix. To contextualize our literature review, we first examine the findings from prior reviews. Then, we present a high-level overview of the general trends from the reviewed studies. Finally, we present the detailed results of the literature review using the image processing pipeline as an analytic framework.

## 6.1 Relevant Prior Literature Reviews

Seven relevant literature reviews were found that discuss vision-based detection and monitoring of: livestock [7,44], wildlife [108,122], or both [1,26,59]. Table 1 summarizes the main focus of each literature review. Two reviews focused on the use of remote sensing imagery captured from satellite, light aircraft, and uninhabited aerial vehicles (UAVs) to support animal science research and wildlife conservation applications. Hollings et al. [59] and Wang et al. [122] reviewed manual, semi-automated, and fully automated approaches that use remote sensing platforms. Both reviews concluded that various factors such as animal size, habitat features like canopy cover, and how visually salient an animal is from its habitat play critical roles in the effectiveness of different imaging platforms and image analysis methods.

For instance, Wang et al. [122] found that pixel thresholding applied to hyperspectral and multispectral satellite imagery was effective for surveying large animals (e.g., elephants) in an open range habitat. Yet, even large animals in satellite images only occupy a few pixels; thus, smaller livestock such as sheep may not be detectable. Both reviews discussed semi-automated approaches that utilize image processing and geographical information system (GIS) software applications [30], such as ArcGIS [79] and eCognition [86]. The object-based image analysis (OBIA) method [52] supported by eCognition enables a user to define rules applicable during image analysis and perform object-based rather than pixel-based analysis. These semi-automated approaches were found to be quite effective in animal classification tasks for estimating wildlife populations. Yet, they require time-consuming user interaction and domain and/or image processing expertise, which renders them unsuitable for PLF applications.

Corcoran et al. [24] reviewed the use of UAV imagery for wildlife and livestock detection. Their analysis of semi-automated and automated image processing methods concluded that choosing the appropriate UAV platform, imaging sensor, and capture conditions can provide acceptably high animal detection rates for certain species and habitats. The use of smaller, lower-flying UAVs and infrared imaging can be effective for detecting small or elusive animals or those living in complex habitats, which are highly challenging conditions for automated methods. DL methods, such as convolutional neural networks (CNNs), were found to be most effective for detecting species with more individual variation and for multi-species classifications. No open challenges for improving computer vision methods were discussed.

Barbedo and Koenigkan [7] investigated the lack of progress for using UAVs to monitor cattle compared to the more prevalent use of UAVs in other farming applications (e.g., crop farming). They identify challenges that hinder the effectiveness and use of UAVs, including aircraft limitations like cost and battery life, contextual factors like airspace regulations, and technical challenges related to image capture and processing. They overviewed some automated animal detection and counting methods, including classic computer vision techniques, ML methods, and emerging DL and transfer learning methods. They concluded that UAV use for livestock monitoring will be inevitable, yet more progress is needed on the identified challenges. In this paper, we specifically explore technical challenges mentioned in their work related to image processing to better understand how to progress the field.

Akbari et al. [1] focused on image and video datasets captured by UAVs and their use in animal detection applications as well as other applications. They characterized the datasets available for each application domain and discussed computer vision tasks for which the imagery has been used (e.g., animal detection, counting, localization). They discussed each dataset's suitability for DL approaches (generally only large datasets). The difficulties of existing image analysis methods were not discussed.

Schneider et al. [108] explored *animal re-identification* image analysis methods based on camera trap data (i.e., imagery obtained from ruggedized cameras installed in wildlife habitats). They found traditional computer vision and ML approaches that used feature engineering based on expert knowledge were primarily being used. Only a few studies were found that utilized DL, including CNNs and Siamese networks (neural networks commonly used to compare images). They argued for the use of DL methods based on their effectiveness in the area of human face re-identification. The complexities of the animal re-identification process were not discussed.

Table 1. Summary of literature reviews between 2018 and 2021.

| Article | Objective |
| --- | --- |
| Akbari et al. [1] | Describes available drone-based datasets used in a variety of computer vision applications. |
| Barbedo & Koeningkan [7] | Explores practical and technical barriers for the limited use of drones in livestock farming. |
| García et al. [44] | Reviews use of machine learning in PLF applications applied to animal health and grazing. |
| Schneider et al. [108] | Reviews use of deep learning in re-identifying wildlife from camera trap data. |
| Hollings et al. [59] | Reviews remote sensing imaging platforms and semi-automated and automated approaches for wildlife and livestock detection. |
| Wang et al. [122] | Details remote sensing imaging platforms and their effectiveness for semi-automated and automated approaches for wildlife detection. |
| Corcoran et al. [26] | Reviews use of drone-based imagery in wildlife and livestock detection. |

Finally, García et al. [44] reviewed the used of ML and DL methods in PLF applications related to grazing and health monitoring. They overviewed methods used for animal behaviour classification, animal monitoring, and animal health applications based on a variety of data sources, including image data. They briefly discussed the use of computer vision and DL methods for animal detection, localization, and individual identification of grazing animals. They found a lack of research in the PLF literature focused on animal comfort, disease diagnosis, or treatment prediction.

Overall, we found no prior reviews that rigorously explore the technical capabilities and challenges of vision-based approaches to outdoor animal monitoring. The above papers all expressed a need for further research on the complexities of computer vision methods. To address this gap, we undertook a new systematic review that identifies current autonomous computer vision methods for this application, examines technical capabilities and limitations for specific animal contexts, and proposes directions for future research to help advance the field for PLF researchers and practitioners.

*6.1.1 High-level Overview of Reviewed Studies*

To help understand current trends in the reviewed study papers, Sankey diagrams were generated that visualize connections between key concepts in the data corpus. Figure 3 represents the prevalence of the different methods and data types used for tasks related to livestock and wildlife research. We classified the domain into three categories where papers focused on: livestock, on wildlife, or both. We labeled these papers based on the methods applied in each study: traditional computer vision, ML, and DL. The third column indicates which type of image data were used. The final column indicates the main goal of the image processing tasks conducted in the papers.

As shown in Figure 2, DL was by far (n=29) the most common image processing method used in the reviewed studies, likely due to recent advances and greater accessibility of DL technologies. The figure also shows that RGB imagery was the most common data type used in the studies, likely because RGB sensors are more affordable, ubiquitous and provide high resolution at reasonable costs. Moreover, many DL models are pre-trained on RGB imagery, and thus have existing knowledge for this imagery type. Animal detection was the most prevalent task found in the data corpus, as detection is usually the first type of object recognition task attempted when applying computer vision algorithms.

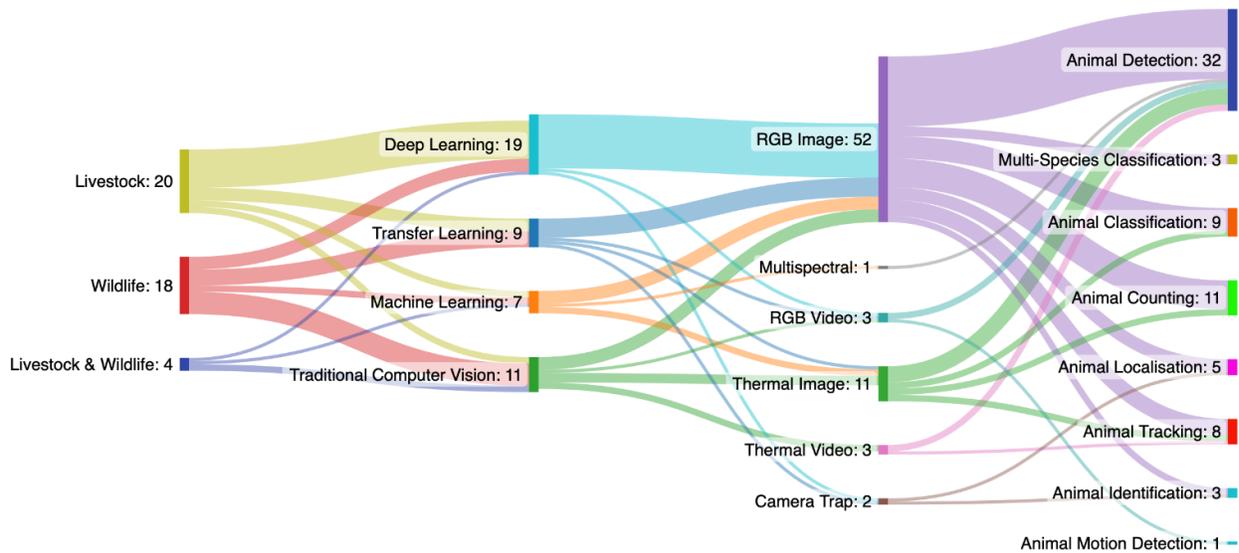

Figure 3. Breakdown of studies in the data corpus by domain, research method, image data, and vision task.

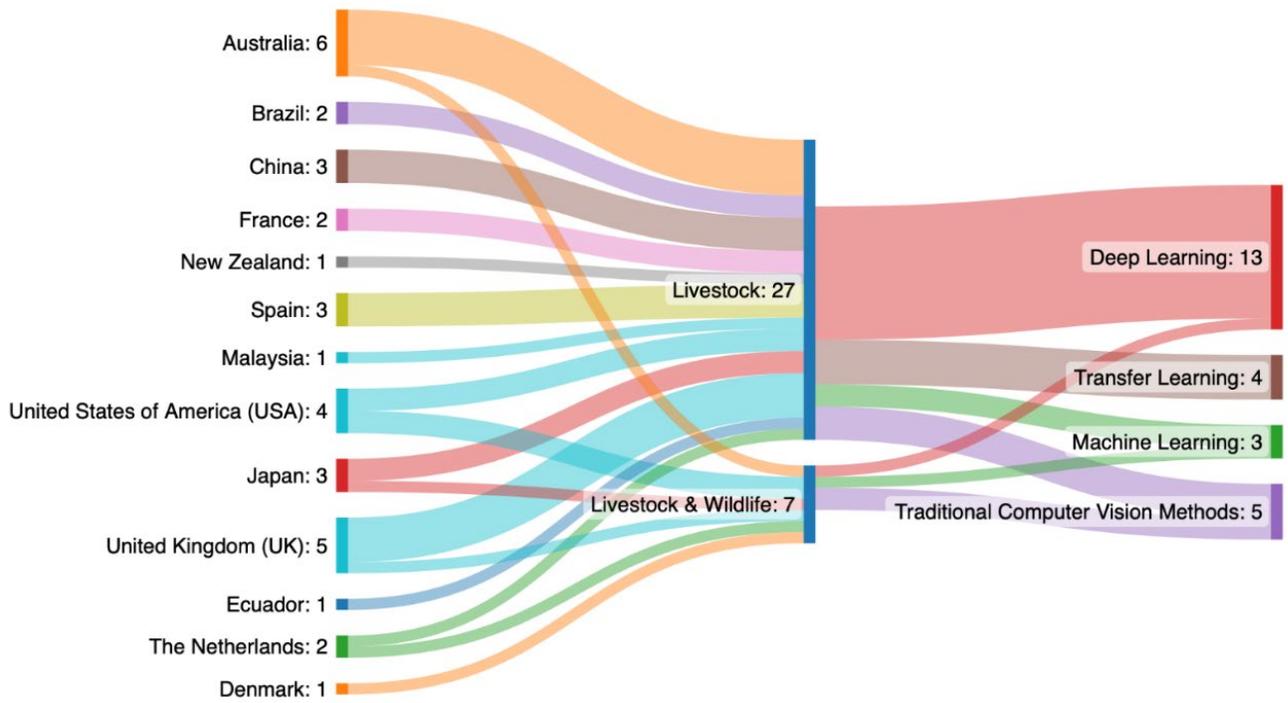

Figure 4. Countries contributing to livestock research and their primary research methods.

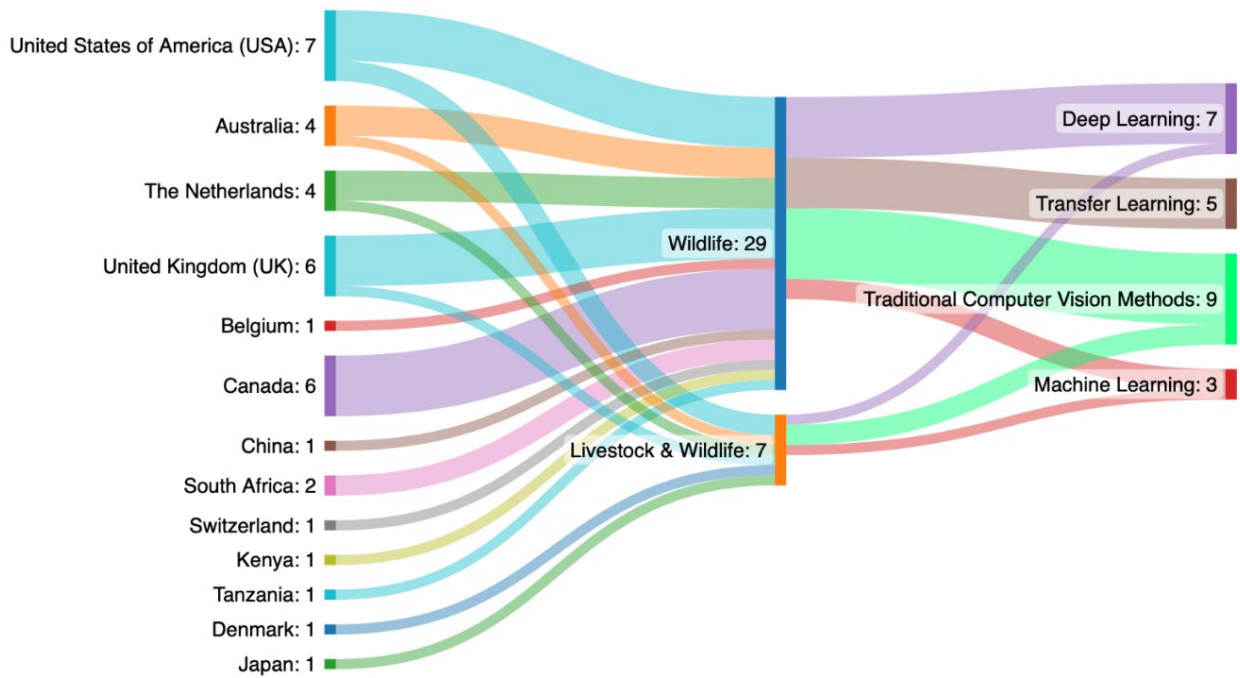

Figure 5. Countries contributing to wildlife research and their primary research methods.

Figures 4 and 5 illustrate which countries are involved in vision-based outdoor monitoring for the livestock and wildlife domains, respectively, and their associated research methods. As shown in Figure 4, Australia tops the list of livestock-focused studies (n=6), followed by the UK (n=5) and USA (n=4) and China, Spain, and Japan (n=3). Australia's strong investment in UAV research and applications [83], and expansive grazing-based livestock farms may explain their strong contributions to this field. The increased rate of meat consumption and livestock production in China and Brazil [82,98] may explain their interest and investment in automating livestock monitoring. The remaining contributions to livestock and livestock/wildlife papers come from countries in Europe, South America, Asia, and the Middle East (n=1 or 2).

A notable omission from the list of livestock-focused studies is India, despite its significant livestock sector, including over 300 M head of cattle and buffalo and over 200 M goats and sheep [111]. A 2020 study found that India had the world's largest cattle and buffalo population (33.4%) and that together with Brazil and China, they held about 65% of the world's cattle and buffalo population [112].

In Figure 4, the authors' home country, Canada, also does not appear on this list. A possible reason is the historic difficulties in gathering datasets of livestock animals using drones. There are a multitude of restrictions on the usage of drones in civilian airspace enforced due to privacy concerns [49] and airspace safety concerns [50]. Our own PLF adoption research has shown Canadian farmers have been slow to adopt PLF technologies in some livestock sectors such as beef production [80], which may further explain the lack of attention focused on automating livestock monitoring. Similar factors may also explain the lack of contributions in this area from African researchers. The lack of papers found from certain countries was also likely impacted by inclusion criteria of English-language papers.

As shown in Figure 5, the USA (n=7) tops the list of contributors to wildlife papers, followed closely by the UK (n=6), Canada (n=6) and The Netherlands (n=5). In Canada, certain wildlife animals (such as polar bears) are held in high regard due to their cultural importance [67] Moreover, Canada and the USA consider wildlife resources a public trust and place a high priority on wildlife conservation [89], likely contributing to the interest in automating wildlife monitoring in both countries. The remaining contributions to wildlife and livestock/wildlife papers come from Australia (n=3) or countries in Asia, Europe, Africa, and the Middle East (n=1 or 2).

These figures show that DL was the primary research method used in livestock (n=13) fields. Traditional computer vision methods were the primary research used in wildlife (n=9) fields, consistent with its prevalence across the whole data corpus. This aligns with recent trends in image processing more broadly.

## 7 DETAILED RESULTS

The studies in the data corpus were analyzed through the lens of the image processing pipeline introduced in section 2 to help understand how well-suited existing computer vision techniques are to address livestock and wildlife domain goals, what contributions researchers have made to address the unique domain challenges, and what challenges remain for developing feasible vision-based animal monitoring. We discuss each pipeline stage in order. However, our review revealed significant overlap between stages in some cases and we also saw various pathways taken through these stages. These variations in image processing workflows are discussed below.

### 7.1 Input Data

Raw data for image processing within the reviewed studies were obtained from either aerial imagery, stationary imagery, or existing datasets. Images were captured using various imaging sensors, such as RGB (green and blue light bands, which are the visible light spectrum) or thermal infrared (a portion of the light spectrum associated with thermal or heat reflectance). The collection of images was compiled into datasets, few of which were made publicly available, and then sent for further processing. We sought to understand the techniques and methods used to obtain these images, and the effects these choices had on their outcome.

*7.1.1 Data Capture*

Aerial imagery was the most common capturing technique used in our data corpus. Aerial imagery typically involved attaching a camera sensor to an aircraft to capture a top-down birds eye view. Most studies (n=19) explored the use of UAVs for facilitating the data capture process. Both rotary and fixed wing UAVs were used to capture images for animal detection and counting. Each UAV platform has advantages and disadvantages [7]. We noted that rotary UAVs were used more often for smaller, contained pastures or environments with unpredictable weather, and fixed winged UAVs were used more in large pastures or open environments.

How far away images are captured from the target can impact how suitable the image data are for certain computer vision algorithms. For instance, UAVs flying at lower altitudes can capture images that show more details of the scene, such as the texture of an animal's skin or fur. Bowley et al. [15] found that CNNs trained on UAV images captured at 75m performed better than those trained at higher altitudes (either 100m or 120m). Christiansen et al. [22] found detecting, classifying, and tracking animals using thermal images was more effective with data captured at closer range (3-10m vs. 10-20m) from a camera affixed to a telescopic boom.

On the other hand, capturing data at a higher altitude allows more area to be captured in a shorter time, which can help address the limited battery life of current UAVs (typically less than 30 mins) [84]. A higher altitude can be balanced with a higher resolution camera, allowing more fine details to be captured. The tradeoff is that higher resolution sensors tend to be heavier and more expensive. Throughout our data corpus, the average resolution for UAV cameras was around 3000x4000 pixels, which is enough to balance out a higher altitude during capturing for many applications [63,93,110].

Typically, aerial imagery captures images from a top-down view. However, Xu et al. [125] captured images from an inclined angle, which showed more of a cattle's profile in the images. This perspective enabled the comparison of head and full appearance detection for cattle in a pasture setting. In the same study, they captured a top-down view of cattle in a feedlot setting, stating that cattle were too clustered and dense to enable effective head profile detection. Thus, the context impacts the suitability of specific camera angles when capturing animals for monitoring purposes.

Stationary imagery was used in some studies to capture side views of animals via cameras placed in fixed positions within the habitat [106,115]. This allowed focus on a precise area; however, the range of coverage was limited. Bonneau et al. [13] overcame this issue by distributing several stationary cameras around a goat pasture, as well as using a time-lapse camera with a wide-angle lens to capture the entire pasture. They proposed a low-cost framework that combined the use of low-cost cameras and ML to detect and monitor the location of individual goats. The authors asserted that using higher resolution imagery would improve their animal detection method.

Several wildlife papers used camera traps. By embedding these cameras in the natural habitat, data can be captured without disrupting the animals [106]. As stationary cameras, they have limited field of view and thus need to be placed according to the pattern of animal movement. Due to the often-remote location in which these cameras are placed, it can be hard and expensive to access and retrieve data [40,115].

Video imagery can be captured by both aerial platforms, or stationery cameras, and provides for a single data file that can be later broken up into multiple frames. Analysis of sequential video frames can provide temporal and contextual information that help improve animal detection, tracking, and classification [3,22]. Video imaging can also help monitor moving animals and track their location. Our review found few studies using video for this purpose, however, as the highly variable nature of outdoor settings introduces significant complexities for existing video analysis methods [73].

*7.1.2Types of Imagery*

The reviewed studies used various image types, allowing researchers to explore the use of different light spectra for different animal monitoring tasks. RGB (red, green, blue) imagery was most commonly used in the reviewed studies, followed by infrared imagery, and then multispectral and panchromatic imagery. Both still and video imagery were used.

RGB (visible spectrum) imagery involves the use of three-band sensors, red, green, and blue, as seen by the human eye [7,21]. It was the most common imagery used for tasks such as animal detection, counting and identification, as it enabled color contrast between the foreground objects and their backgrounds. For example, Andrew et al. [3] used an ideal RGB dataset consisting of Holstein Friesian (black and white) cattle against a green pasture for animal detection.

Capture conditions, such as illumination, weather, time, camera errors, and habitat conditions, can decrease contrast and make detection challenging [7]. However, RGB imagery was shown in the reviewed studies to be beneficial for mitigating these issues, allowing the use of pre-processing methods such as color and contrast manipulation (see section 6.2) to enhance image quality [8,15]. Moreover, when higher resolution images are captured, significant information can be provided by RBG imagery for analysis [118].

Infrared (thermal) imagery was used effectively in some studies for detecting and counting animals as it enables the separation of warm-blooded animals from their backgrounds via thresholding [22]. Unlike other image types, thermal imagery penetrates some obstructions, such as foliage and canopy cover [7,48]. It was also used to track wildlife in low light conditions [77].

A major disadvantage of thermal imagery is its lower spatial resolution compared to other image types [118]. This can lead to a uniform heat distribution at higher altitudes and makes object recognition nearly impossible. However, this can be mitigated by capturing data at lower altitudes [22]. Insufficient contrast between animals and their surroundings in thermal imagery can also be caused by the sun heating objects in the environment. Capturing data in early morning before too much sun exposure can help avoid this issue [7,48]. Some studies used infrared imagery to filter out false positives caused by objects heated by the sun as well as segmentation of herds of animals into individuals [22,25,77,123].

Multispectral imagery is unique as it captures a variety of wavelengths. It has been largely used in agricultural applications to detect vegetation rather than animals [7]. Chrétien et al. [21] explored the use of four combinations of light bands to detect and count white-tailed deer in an outdoor setting. Other studies have explored combining multispectral and panchromatic imagery for animal detection [67].

Exploratory studies by Terletzky et al. [118] suggest that multispectral imagery could be used to distinguish between different species or breeds of animals with different coat types or colors. They used a spectrometer to capture different light spectra of different animal species (horse, cattle, elk), and found significant differences between brown and black cattle within the red and near infrared bands. However, elk were found to have high within-species variations for all spectral bands except blue, making it difficult to reliably detect them from other species.

*7.1.3 Other data sources*

Training ML and DL models require large image datasets. Yet, the data corpus shows a lack of publicly accessible datasets with relevant or sufficient livestock or wildlife content. Therefore, many studies resorted to capturing their own data to create custom datasets, as described above. Researchers also used data augmentation (generating new images from existing ones) by applying various image manipulation techniques such as rotations, vertical/horizontal shifts and flips, brightness adjustments, zooming in and out, mosaicking/stapling overlapping images, and introduction of random noise on the original images [63,91,106]. Data augmentation can increase the number of input images or produce images with desired characteristics to improve the accuracy and robustness of DL models [107].

**7.2 Pre-Processing**

The reviewed studies used various pre-processing techniques to modify imagery to increase the success of image analysis algorithms. For DL methods, this stage often involved annotating and labeling metadata for model training and testing.

Pre-processing methods were used to enhance raw images, to highlight key features, and/or to improve the quality of the input data. Poor illumination and other environmental factors can result in low contrast images, making it difficult to distinguish an animal from its background [93]. Contrast enhancements were applied in some studies to exaggerate pixel differences, making edges clearer and better distinguishing objects from their backgrounds [4,8,9,48]

Color correction was also used to improve image characteristics for later processing, such as correcting colors due to equipment errors (e.g., a blue sensor failing) or converting images to greyscale to shorten computation costs [8,103,105]. Barbedo et al. [8] converted their images to the CMYK (cyan, magenta, yellow, black) color spectrum, increasing the saliency of certain image features while decreasing others to improve animal detection.

Thresholding was not commonly used as a standalone pre-processing technique. Instead, it was often combined with other pre-processing methods, such as edge detection methods that highlight features and reduce noise. Vayssade et al. [121] applied thresholding to improve the detection and tracking of goats from RGB imagery. Because red goats were indistinguishable from certain background features (e.g., dry weeds), they applied thresholding to images in multiple color spaces (RGB, HSV, YCrCb, HSL, and LAB), each of which revealed different aspects of the scene. Several studies applied thresholding to thermal imagery captured from UAVs as a first step to animal detection using traditional computer vision methods [22,48,77].

Other image enhancement methods included Gaussian smoothing, blurring, or applying filters to reduce image noise and eliminate the background to focus on the object of interest [4,64,121]. Chrétien et al. [21] applied polynomial equations to correct distortions introduced by moving deer within their UAV imagery. Image pixels were matched to corresponding ground coordinates to facilitate image correction.

Methods were also used to modify the image format to meet requirements of certain algorithms, including mosaicking, cropping, scaling, and image transformations, such as rotations. Mosaicking combines images together, forming one or more larger images to be examined [15,21,93]. Georeferenced ground targets and reference mosaics from prior studies were sometimes used to facilitate image mosaicking [21]. Mosaicking was also used together with cropping to prevent overlap and mitigate the effects of overcounting and undercounting [15,21].

Cropping was commonly used to create images of uniform size to meet algorithm requirements [4,93,105,126]. For instance, in a CNN-based DL model, each pixel in a uniformly sized image maps onto an exact input node in the model. Cropping often involved removal or adjustment of the outer edges of an image to remove background areas around the animals [4,108,128].

Other image transformation methods, such as rotating or flipping, were rarely used for pre-processing in the data corpus. During capture, Andrew et al. [3] found that moving cows caused an imbalance between images with and without cattle in their dataset. Therefore, they implemented data augmentation, including rotation of images, to balance out their dataset. Kellenberger et al. [63] applied random rotations to their images to make their algorithm more robust. Rahnemoonfar et al. [93] used random rotating, random flipping, and cropping when training their algorithm.

Labeling whole or parts of images based on user-defined semantic categories was commonly used during pre-processing, for instance, labeling images to denote whether or not they contained a target animal or not, or the species of the animal [9,14,45,105,115]. Such labels were necessary to provide the supervised ML or DL models feedback during model training. Tools such as the freeware, LabelMe, were often used to perform image labeling [125,126].

### 7.3 Selecting Areas of Interest

This stage of the pipeline involves the identification of the most significant regions within an image, for instance, delineating between animal properties and static backgrounds. Studies within the data corpus analyzed still images, video data, or both.

Many studies created feature maps that represented a region of interest generated by extracting animal features and allowing their animal detection algorithms to focus on the data within that area. Principal Component Analysis [61] was used in some studies to analyze images for distinguishable animal features to highlight areas of interest [21,117]

The histogram of oriented gradients (HOG) method [29] was used in some studies to highlight animal features within an image [22,77,119,120]. Torney et al. [119] and Valletta et al. [120] used a rotation invariant version of HOG to extract distinct wildebeest features from UAV imagery because animals were at different orientations in the dataset.

Okafor et al. [88] investigated various feature extraction methods, including HOG, color histograms, and both, to select cattle features from greyscale UAV imagery that were then used as input to ML algorithms. They found these methods performed worse than a DL model applied to the same dataset (see Section 6.4.3).

Thresholding was used in various studies to create clusters of pixels to generate feature maps, bounding boxes, or binary masks of an image [22,121,123]. When analyzing thermal imagery, researchers commonly set a threshold to a certain pixel value and ignored all areas of the image above or below that value to help separate "warm" bodies from "cooler" surroundings [22,48,123]. Sarwar et al. [105] used thresholding on images to determine regions of interest of RGB images that contained sheep because the sheep were expected to be the brightest objects in the frame. Similarly, Christiansen et al. [22] detected regions of interest by applying thresholds to each image frame, followed by the extraction of animal features using HOG.

Binary masks were used to improve classification by eliminating irrelevant areas of the image to improve object detection algorithms. Studies commonly used thresholding or feature extraction techniques to create binary masks of images [8,25,38,48,87,121]. Oishi and Matsunaga [87] used thresholding techniques to create binary masks using Laplacian histograms [124], the P-tile method [116], and the Otsu method [90]. Binary masking was also used to track animals in adjacent video frames [48]

Barbedo et al. [8] used a unique approach to selecting areas of interest. They first processed their image using a CNN to generate regions of interest. These image segments were then divided into four squares that either contained animal features or not. Binary masks were then created for each square using color manipulation, thresholding, and mathematical morphology. They explained that binary masking was applied *after* rather than *before* the data were input to their CNN because it handled variations in contrast and illumination well and could provide a rough estimate of an animal's location within the image. This helped focus the additional image processing steps, which were needed to mitigate the shortcomings of their DL model.

Some studies generated bounding boxes for highlighting animals in the imagery by using CNNs [3,4,25]. Other studies used methods such as thresholding or binary masking to locate an area of interest and then to draw bounding boxes around the detected animal [38,48].

Fang et al. [38] used the optical flow method to identify apparent motion of pixels in successive images. They applied thresholding to generate binary masks and to draw bounding boxes around animals that appeared to be moving. Generating the binary masks helped the authors separate animals from the background, improving the accuracy of the bounding boxes. They noted their technique worked best for animals closer to the camera, because the image had more pixels representing the animal.

### 7.4 Processing Selected Areas of Interest

The reviewed studies used a number of techniques to process selected areas of interest, including traditional computer vision techniques, ML techniques, and DL techniques. There was a clear trend in recent studies toward DL, leveraging the power of these approaches to improve image processing. Animal detection typically occurred at this stage, either using feature analysis algorithms, ML or DL models. Bounding boxes were often applied to detected animals. Depending on the image processing goal, the process either stopped (i.e., only animal detection and/or counting were performed), or images were sent to the last stage for further analysis.

*7.4.1 Traditional Computer Vision Techniques*

Most studies using traditional computer vision techniques appeared early in our data corpus. These techniques tended to leverage expert knowledge to create customized algorithms highly tailored to a specific animal species and to potential distractor objects to be ignored (e.g., structures, vegetation). Thus, these algorithms are difficult to transfer to other species or environments. For example, Lhoest et al. [70] knowledge of hippo anatomy to determine reasonable distances and relative angles between two proximal areas of interest of a single, partially submerged hippo (i.e., one area representing the head and the other representing the rear end) in thermal imagery. Oishi and Matsunaga [87] used the average length, height, and width of cattle to create a species model for their detection of moving wild animals (DWA) algorithm. This model was used together with known data capture information (e.g., flight altitude, imaging angle) from UAV imagery to estimate the expected size of an animal in the data. Sequential frames of time-lapsed images were analyzed using this information and then potential candidates in adjacent frames were compared to detect moving animals. This approach eliminated stationary objects from consideration, including animals at rest.

Studies involving thermal imagery typically used traditional computer vision techniques and/or ML techniques due to their lower resolution and limited available color information that can render DL models less effective. Gonzalez et al. [48] used the template matching binary mask (TMBM) technique to detect koalas from thermal imagery. Reference image templates of koalas were used to search for matches in the image data and label them. Multiple templates were used to represent different sizes and postures to improve performance. This technique was combined with analysis of consecutive video frames to increase the reliability of detection by only counting animals that were detected in 10 consecutive frames.

Sarwar et al. [105] compared a traditional computer vision technique to several DL mehods for livestock detection. They used a custom blob detection algorithm [114] to detect sheep. Blobs smaller than a certain area threshold were discarded, while blobs above (roughly) twice the mean blob area were counted as two sheep. The blob detection algorithm was compared to several R-CNN models and was found to be more accurate (95.6% precision; 99.5% recall) than the best of the tested R-CNNs (~95% precision; ~83% recall). However, this blob detection approach may not be suitable to other environments that contain potential distractor objects of similar color and size as the target animal. The algorithm could also fail if the size of the target animals varies considerably or if animals of various ages are present in the data.

*7.4.2 Machine Learning*

The data corpus shows a shift away from traditional, handcrafted animal detection algorithms towards the use of ML object classifiers. Studies primarily used supervised ML approaches, which rely on pre-classified image data to train models to learn characteristic features that represent a specific class of animals or animal behavior. Many ML object detection methods have been developed to address problems in various domains. Yet, with no established best practices for wildlife and livestock detection tasks, the data corpus shows researchers exploring a wide variety of ML approaches with varied results.

Van Gemert et al. [45] compared two ML methods for detecting cattle from RGB UAV video imagery: the Deformable-Parts Model (DPM) approach [39] (default greyscale approach and RBG color variants) and the ensemble of exemplar Support Vector Machine (SVM) approach [81]. The exemplar SVM approach performed better animal detection (precision: 66%, recall: 72%) than the DPM approaches (precision: ≤ 30%, recall: ~40%). They concluded that the small size of cattle in the UAV imagery was unsuitable for the DPM approach, due to the lack of discernable features to enable the algorithm to recognize cattle by their connected parts.

Longmore et al. [77] used an SVM classifier [27] with inputs from a HOG filter for detecting cattle from thermal UAV imagery, yielding only moderate results of 70% accuracy for animal detection. This method may have also suffered from the small scale of animals visible in the UAV imagery and lack of discernable features. Moreover, HOG filters are sensitive to rotation, and thus, the approach may have been negatively impacted by cattle positioned at different orientations given the birds-eye view in the UAV imagery. To address this issue, Torney et al. [119] used a rotation-invariant HOG filter [76] as input to an AdaBoost object classifier [43] to analyze RGB aerial imagery. Their ML approach yielded slightly better animal detection performance (precision: 74%, recall: 85%). Valletta et al. [120] used a rotation-invariant HOG filter as input to a Random Forest classifier [16] for multi-species detection, with results differing across species. Zebra detection performed the highest (precision: 100%, recall: 100%), while wildebeest detection produced many false positives and false negatives, lowering performance (precision: 92%, recall: 86%). This was most probably caused by the difference in observable patterns on the animals hide.

Vayssade et al. [121] compared the performance of ten different ML classifiers for detecting goats and their activity (see section 6.5). They used several different image features, including HOG, as inputs to the classifiers. The Linear Discriminate Analysis (LDA) method [100] performed the best for detecting goats (accuracy: 96.8%) with Random Forest providing similar detection accuracy. They explored LDA in detail and found it produced many false positives but fewer, though still substantial, false negatives. Overall, it led to low precision (45%) and only moderate recall (74%) scores.

Various ML object classifiers have been explored to detect and count deer. Christiansen et al. [22] analyzed RGB and thermal imagery of deer taken from close (3-10m) and medium (10-20m) range using the k-Nearest Neighbor (kNN) method [12]. Combining the RGB and thermal imagery during the analysis yielded reasonable detection performance for the close range (precision: 95%, recall: 86%) that was improved by adding a simple tracking algorithm across sequential video frames (precision: 96%, recall: 91%). Chrétien et al. [21] analyzed RGB and thermal UAV imagery using the Maximum Likelihood method [101]. Their best results were obtained by analyzing both the RGB and thermal imagery, but this approach still achieved poor detection performance (66% accuracy). RBG imagery and thermal imagery alone each produced a substantial number of false positives during analysis.

Chrétien et al. [21] also used an unsupervised ML approach, the k-Means method [78]. This approach has the advantage of not requiring labelled data for model development. It produced strong animal detection performance (it detected 100% of the detectable deer in the images); however, it yielded a substantial number of false positives, especially with RGB imagery alone. Terletzky et al. [117] used an unsupervised ML approach, the Iterative Self-Organizing DATA (ISODATA) analysis technique algorithm [5], to detect and count cattle and horses. They obtained modest results (82% accuracy), but this approach also yielded substantial false positives, leading to significant overcounting.

### 7.4.3 Deep Learning

Following a broader trend in computer vision, livestock and wildlife researchers have largely abandoned traditional ML approaches in favour of Convolutional Neural Network (CNN)-based supervised learning approaches for animal detection and related tasks. CNNs are artificial neural networks that perform pixel-based analysis on images by using multiple layers of connected nodes that extract learned features, often iteratively, for localization and classification of objects [68]. Table 1 is a summary of DL studies from the data corpus and their reported results. Initial studies explored custom built CNN models for detection and counting of livestock [20,88,99] and wildlife [14,88], yielding promising results.

However, high numbers of false positives were also commonly reported [14,20]. Moreover, some studies used unrealistic training data that were highly curated and enhanced for proof-of-concept purposes. For instance, Okafor et al.'s [88] training and test dataset contained images of cattle that contained a single, full cow with extraneous background areas cropped out. More realistic data caused challenges for these early CNN models, including misclassifications – including high numbers of false positives – and missed small or clustered animals [20,63,99].

Several studies successfully reduced false positive detections for their models by applying hard negative mining [127] during model training [15,63,91]. This training technique provides feedback to the model for gross misclassification errors and retrains it on misclassified data until performance improves [127]

Despite the promising results of CNNs, they also introduce several challenges for animal detection. First, they require very large datasets (i.e., millions of images) to sufficiently train robust models. Such datasets do not currently exist in the livestock field, and few exist in the wildlife domain. Image datasets also need associated labels to facilitate feature learning during training. Another challenge is the size of newer CNN models. Recent CNN architectures have dozens to hundreds of layers optimized for image processing that have led to significant improvements in object detection and related tasks [130]. These large models, referred to as DL models, are extremely computationally expensive and time consuming to train. Recent papers in the data corpus addressed the above issues by employing pre-trained DL models.

Pre-trained models are already trained on massive datasets that contain many classes of everyday objects, including different types of animals, to provide a base level of knowledge in the models. Some reviewed studies employed transfer learning, a method whereby a pre-trained model is further trained with a smaller, domain-specific dataset to refine its knowledge base. Most studies employing pre-trained models or transfer learning pre-trained their models using subsets of or the whole ImageNet dataset [102]. Thus, for brevity, we omit the training dataset below, unless another dataset was used.

Barbedo et al. [9] and Schneider et al. [106] explored the potential of pre-trained DL models for livestock and wildlife research. Barbedo et al. [9] compared the performance of 15 CNN architectures for detecting cattle, with NasNet Large [131] and Inception [113] performing the best on their RGB UAV images. Schneider et al. [106] compared the performance of six DL models for multi-species detection from RGB wildlife camera trap images. Additionally, they included an *ensemble* model in which all models vote on the classification to produce a result. They also tested the models on data from an untrained camera trap location, leading to reduced performance across all models.

Researchers also used DL models as one component of larger workflows to improve animal detection. To mitigate several limitations of CNNs Barbedo et al. found in their aforementioned 15-model comparison, they proposed an alternative workflow. They first used a CNN (NasNetLarge) to identify potential cattle and then applied a series of more traditional pre-processing algorithms, as discussed in section 6.3, to help differentiate between clustered cattle and to detect calves [8]. Their workflow yielded strong performance for cattle detection (precision and recall: ~97%).

Kellenberger [64] adapted the AlexNet DL architecture [65,66] to include two branches, one for animal detection and one for animal localization, and to perform multi-species detection on RGB UAV imagery of African wildlife. Compared to a state-of-the-art CNN variant, R-CNN (Region Proposal CNN) [47], their model produced far fewer false positives, increasing precision by 25%, and it was also faster because fewer regions were considered.

In a follow-up study Kellenberger [62] used a CNN based on the ResNet-18 architecture [54] together with the optimal transport (OT) domain adaptation method [28] to address the brittleness of DL models applied to untrained domains or data capture locations. The OT method is a general statistical method used to map probabilistic distributions in one domain to probabilistic distributions in another. The OT approach helped improve precision and reduce false positives during multi-species detection compared to the CNN alone. Yet, both approaches yielded relatively low performance (precision: <30%).

Eikelboom et al. [35] and Tabak et al. [115] used variations of the ResNet CNN architecture (ResNet50 and ResNet18, respectively) for multi-species wildlife detection. Reasonable species detection was found in both studies; however, performance varied across species (reported accuracies: 80-97% for the same locations where the training data was collected, 31-91% for different locations). Eikelboom et al. [35] reported large numbers of false positives, possibly due to canopy cover in the UAV imagery being processed. To help improve the multi-species detection process, Tabak et al. [115] also proposed a binary classification CNN (based on ResNet18) that could be used to detect and filter out "empty" (no animal present) images first before sending remaining "animal" images to the multi-species CNN classifier. The "empty" classifier yielded 97% accuracy on the camera trap datasets tested.

Several researchers used the You Look Only Once (YOLO) [96] DL model and variants to achieve fast detection. Corcoran et al. [25] used YOLO, together with Faster R-CNN [97], to detect koalas from UAV thermal imagery. They used the bounding boxes produced by both models to produce binary masks of likely animals. They then applied a greedy tracking algorithm [51] to successive video frames to find consistent detection of animals across several video frames (precision: 49%). The fairly low detection rate for this study is not surprising given the canopy cover in the koala habitat, the relatively low pixel resolution of thermal imagery compared to RGB imagery, and the fact that pretrained networks tend to be trained with RGB imagery.

Shao et al. [110] used YOLOv2 [95] for detection and counting of cattle from RGB UAV imagery. To improve performance, they merged the detection results with a 3-dimensional (3D) reconstruction of the scene. They obtained reasonable results with models tested on data collected from the same location (precision: 96%, recall: 95%), but much lower performance on data collected from a different location (precision: 77%, recall: 66%). Their 3D reconstruction technique missed animals moving quickly and those not moving.

Bonneau et al. [13] successfully used tinyYOLOv3 [94], pretrained on the Pascal VOC dataset [37] to detect goats from RGB imagery captured from a stationary time-lapse camera (precision: 90%, recall: 84.5%). Bounding boxes produced by the model were then used to determine the locations of the goats in their pens.

R-CNN and variants were used by Sarwar et al. [105] and Xu et al. [126] for sheep and cattle detection from RGB UAV imagery. Both studies achieved high detection performance (accuracy/precision: 96+%). Xu et al. employed a Mask R-CNN model [53] that produced a contoured mask around the animal rather than the bounding box produced by other DL models. The authors mentioned that such mask could be used to potentially determine behaviour or health states.

Faster R-CNN [97] pretrained on VGG-M1024 [85], was used by Andrew et al. [3] to detect regions of interest in a larger workflow to identify individual Holstein Friesian cows from RGB UAV imagery. This part of the process achieved near perfect detection performance (precision and recall: 99%). The bounding boxes produced by Faster R-CNN were then passed to the next stage for further processing.

Some studies explored simpler ANN models to improve training or processing speed for animal detection applications. Rahnemoonfar et al. [93] designed a custom CNN, called DisCountNet, with 16 layers to select regions of interest and then count animals inside the selected regions. Their approach was similar to the aforementioned approach by Tabak et al. [115] that first filters out images with no animals. However, in DisCountNet, the portion of the model that selects regions of interest is only a few layers compared to the full DL model proposed by Tabak et al., making their filtering process very fast. DisCountNet was quicker to train than the RetinaNet [75] and the CSRNet [72] DL models, processed images in real-time (34 frames per second) and outperformed the DL models on detection and counting metrics.

Two studies by Sadgrove et al. [103,104] used an alternative neural network approach, called Extreme Learning Machine (ELM), to detect cattle from RGB imagery from a stationary camera. ELMs use a special training process that assigns random weights for input nodes and does not use the popular backpropagation algorithm used in CNN training [31]. Their studies explored the potential of using different color spaces (RGB, HSV, Y'UL) as input to an ELM for animal detection. They achieved only moderate detection performance compared to other CNN approaches discussed above, with their performance reported as 86% accurate in 2017 and 85% precision / 91% recall in 2018s.

Only one reviewed study focused on motion detection of animals. Liao et al. [73] proposed a variation of the FlowNet2 CNN, for estimating optical flow between sequential video frames to detect motion. Their model, FlowNet2-IAER, introduced custom illumination adjustments and edge refinement to improve motion detection. Their proposed model was 9.14% more accurate for detecting contours and edges across frames than the original FlowNet2 model. The authors discuss the challenge of measuring accuracy for motion detection algorithms as little ground truth data exists.

Table 2. Summary of studies using deep learning models for wildlife and livestock animal monitoring tasks. Note, precision, recall, and F1 scores metrics are reported if available / possible to calculate. Otherwise, reported performance measures are provided.

| Article | Model | Reported Performance (Measures from article reported) | Animal Type | Task |
|---|---|---|---|---|
| Andrew et al. 2017 | VGG-M 1024 end-to-end RCNN | Detection & Localisation: 99.3 mAP†     Identification & Localisation: 86.07% mAP | Cattle | Identification, localisation, tracking |
| Andrew et al. 2019 | YOLOv2-based species detector & InceptionV3-based RCNN | Combined Detection & Identification Accuracy: 91.9% | Cattle | Detection, identification, localisation |
| Barbedo et al. 2019 | Top 2 CNNs: NASNet Large Xception | CNN     NASNet Large     Xception<br>Precision     99.3%     96.8%<br>Recall     99.3%     96.8%<br>F1 Score     99.5%     96.8% | Cattle | Detection |
| Barbedo et al. 2020 | NasNet Large | Overall precision, recall and F1-score: 97.4% | Cattle | Classification, counting, detection |
| Bonneau et al. 2020 | YOLO | Precision: 90%     Sensitivity: 84.5% | Goats | Detection, localisation, tracking |
| Bowley et al. 2017 | Custom-built CNN | CNN     Expert trained     Matched trained     Unmatched trained<br>Accuracy:     99.81%     99.76%     99.64% | Wildlife | Animal detection |
| Bowley et al. 2019 | Custom-built CNN | Reduced errors on citizen scientist training data from +150% to -3.93% & expert data from +88% to +5.24%. Lowest altitude provided best predictions of +11.46%. Aggregate data provided best results of -3.93% | Wildlife | Animal detection |
| Chamoso et al. 2014 | CNN | Cattle Accuracy: 91.8% | Cattle | Animal detection; counting |
| Corcoran et al. 2019 | Faster-RCNN & YOLO used together | Precision: 49% | Koalas | Detection |
| Eikelboom et al. 2019 | RetinaNet | Animal     Elephants     Giraffes     Zebras<br>Accuracy:     95%     91%     90% | Elephants, giraffes, zebras | Animal detection, multi-species classification |
| Kellenberger et al. 2018a | CNN based on ResNet-18 | Modified method increased precision versus unmodified CNN | Wildlife | Animal detection |
| Kellenberger et al. 2018b | CNN based on ResNet-18 | Recall     Precision     F1<br>70%     50%     60%<br>80%     30%     40% | Wildlife | Animal detection |
| Kellenberger et al. 2017 | Faster R-CNN CNN model | Precision: 60%<br>Recall: 74%<br>F1 Score: 66% | Wildlife | Animal detection, localisation |
| Liao et al. 2021 | FlowNet2 FlowNet2-IAER | FlowNet2 F1 Score: 4.85%<br>FlowNet2-IAER Score: 4.02% | Cattle | Animal motion detection |
| Okafor et al. 2018 | original CNN Fine-tuned CNN | Original CNN best test accuracy: 99.41%<br>Fine-tuned best test accuracy: 99.65% | Cattle, birds, fish | Animal detection |
| Peng et al. 2020 | Faster R-CNN ResNet-101 | F1 Score: 90% | Kiang | Animal detection |
| Rahnemoonfar et al. 2019 | DisCountNet RetinaNet CSRNet | mAE (mean absolute error):<br>CSRNet: 1.58<br>RetinaNet: 1.24<br>DisCountNet: 1.65     mSE (mean squared error):<br>CSRNet: 4.49<br>RetinaNet: 3.54<br>DisCountNet: 4.98 | Livestock and wildlife | Animal detection; counting |
| Rivas et al. 2018 | CNN | Accuracy: 95.5% | Cattle | Animal detection; counting |
| Sadgrove et al. 2017 | Color Feature ELM | Accuracy: 86% | Cattle | Animal detection |

Table 2 (cont'd).

| Article | Model | Reported Performance (Measures from article reported) | | | | Animal Type | Task |
|---|---|---|---|---|---|---|---|
| Sadgrove et al. 2018 | Color Feature ELM | Precision: 78% Recall: 93% | | | | Cattle | Animal detection |
| Sarwar et al. 2018 | R-CNN | Precision: 97% Recall: 87% | | | | Sheep | Animal detection |
| Schneider et al. 2020a | DenseNet201 (best) | Accuracy on trained locations: 95.6% F1 score on trained locations: 79.4% Accuracy on untrained locations: 68.7% F1 score on untrained locations: 69.8% | | | | Wildlife | Multi-species detection |
| Schneider et al. 2020b | ResNet152: overall best. DenseNet201: best for mAP@1 for tiger dataset | Tiger Dataset for Siamese Similarity Learning Model: mAP@1: 80.3%, mAP@5: 87.7% | | Tiger Dataset for Triplet Loss Similarity Learning Model: mAP@1: 86.3%, mAP@5: 99.6% | | Wildlife | Identification |
| Shao et al. 2020 | YOLOv2 | Dataset: Dataset 1 Dataset 2 | Precision 95.7% 77.4% | Recall 94.6% 66.1% | F1 95.2% 71.3% | Cattle | Animal detection; counting |
| Tabak et al. 2020 | CNN | Species model accuracy: 96.8% Empty animal model: 97.3% | | | | Wildlife | Animal detection; multi-species identification |
| Xu et al. 2020a | Mask R-CNN ResNet-101 | Classification accuracy: 96% Livestock precision: 95.5% Cattle Precision: 95.4% Sheep Precision: 96% Recall: 95% | | | | Cattle and sheep | Animal detection; counting |
| Xu et al. 2020b | Mask R-CNN ResNet-101 | Full appearance detection: 96% Head detection in pasture: 92% Full appearance detection in feedlot: 94% | | | | Cattle | Animal detection |

†maP = mean average precision

## 7.5 Decision Making

This stage represents the culmination and final output of the image processing pipeline. During the decision-making stage, researchers decide how to interpret their data and make decisions based on the output of the prior stages. A decision can range from basic (e.g., is the target object (a face, etc.) in the image?) to more sophisticated (e.g., what emotion is being expressed on the detected face). In the reviewed articles, most decisions were basic (animal detection, counting, multi-species classification, tracking, or motion detection). Processing largely concluded at the prior stage with little to no further analysis. In some cases, researchers applied late-stage techniques to enhance their results and improve decision outcomes. Only a few papers went beyond these basic detection tasks to perform, for instance, individual animal identification [2] and behavioural classification [121]. We overview challenges related to performing the basic decision tasks with confidence. Then, we describe the few methods found in the data corpus for making more sophisticated decisions.

As discussed in section 5 (see Figure 3), animal detection and counting were by far the most common decision tasks in the reviewed studies. Using detection results to inform counting decisions seems like a straightforward task, but the studies highlighted several challenges for doing this effectively. One challenge is knowing whether detected animals are unique and whether they represent all animals in the area. Surveying animals using aerial imagery can lead to missed animals or animals appearing more than once because both the aircraft and the animals can move during data capture, and imagery taken of nearby regions can contain overlapping footage. A common strategy to deal with double counting were to mosaic images together to create a complete scene [15]. Shao et al. [110] applied a 3D surface reconstruction technique to merge data over consecutive video frames from UAV data and to eliminate duplicate detections from adjacent scenes.

Environmental and capture issues and crowding also posed challenges for animal detection algorithms. Barbedo et al. [8] found they undercounted cows in their images when calves were present because they tend to stay close to their mothers and are harder to detect due to their small size. Xu et al. [126] noted their model performed better when ten or more cattle were in an image because there were fewer misclassified non-animal objects.

Assessing the performance of detection, counting, and multi-species classification methods requires ground truth data to evaluate model results. In livestock contexts, this information was largely known because datasets were typically created from captive livestock herds, where the number and type of livestock were known [8,20,99,110,126] However, in wildlife settings, populations are typically unknown. Thus, alternative methods were needed to determine ground truth, such as manual image analysis by human experts [35,67,91,119] or crowdsourcing [64].

To improve counting accuracy (and detection performance), some studies further processed initial detections. Such strategies included only counting animals that were detected in multiple sequential frames of video data [48], using object tracking methods [13,64], or further processing the bounding box regions output from a CNN animal detector using additional image processing techniques (e.g., color manipulation and applying mathematical morphology) to separate clustered animals [8].

Tracking individual animals over a series of images was used to detect animals and their movements [3,13,22,48,121]. Christiansen et al. [22] tracked animals by creating mathematical predictions of their next movements based on previous movements and then checking if the animal was in the predicted zone. Fang et al. [38] used the optical flow analysis of an image to identify and track moving zebras based on the relative motion difference between the foreground and background of an image.

Only two studies focused on identifying individual animals. Andrew et al. [3] used tracking algorithms and DL to identify cattle in a multi-stage process. As aforementioned, they used a R-CNN model to detect cattle. Resultant bounding boxes for sequential video frames were sent to a Kernelized Correlation Filter (KCF) tracking algorithm [56]. Tracked regions were then input into a Long-term Recurrent Convolutional Network (LRCN) [32] that contained a CNN (Inceptionv3 [113]) with Long Short-Term Memory network (LSTM) [57] layers that performed the individual animal identification (accuracy: 98%).

Schneider et al. [107] studied two similarity learning networks: the Siamese network [17] and the Triplet Loss network [58], which compare input data to known samples of target individuals to determine the degree of similarity and probability of a match. Both models process data through parallel CNNs before similarity measures are applied. For each similarity network, they considered five different CNN models: AlexNet, VGG19, DenseNet201, ResNet152, and InceptionNet v3, and five different species datasets (human face, chimpanzee face, whale, fruit fly, and Siberian tiger). The Triplet Loss network (using ResNet152) performed the best across the five species (mAP@5: 94.5%).

Only one paper in the data corpus conducted a behavioural analysis on detected animals. Vayssade et al. [121] compared the performance of ten different ML classifiers to determine the activity (sleeping or eating) of goats. Using a number of shape features as input to the ML models, they found the AdaBoost [42] classifier yielded the best results, with "eating" being correctly detected 78% of the time and "sleeping" 92% of the time.

## 8 DISCUSSION

The typical goal of precision livestock monitoring approaches is to "precisely" monitor individual animals to detect health and welfare issues as early as possible and improve animal outcomes and farm productivity [11]. This review found that vision-based approaches for monitoring livestock in outdoor environments can detect and count animals fairly reliably in many contexts. However, more research is needed on detecting animal state and behaviour to achieve the goal of livestock health and welfare monitoring. Specific challenges must be overcome. In this section, we reflect on the implications of current capabilities for livestock farming contexts and discuss the open challenges to improve current capabilities. We then discuss the potential to move beyond basic detection capabilities to enable feasible vision-based technology solutions for livestock monitoring in outdoor habitats in the future.

### 8.1 Current Capabilities Related to PLF

We intentionally cast a wide net when selecting papers to include in our data corpus to understand the current capabilities and trends of vision-based approaches for monitoring animals in an outdoor setting, including papers outside the field of agriculture. Our hope was to uncover techniques from a broad spectrum of research with similar goals that may be applicable to the PLF context. However, the end-goal of some of the surveyed research differed from our PLF context. Therefore, it is worth discussing the feasibility of the vision-based monitoring capabilities we found to support or be adapted to PLF goals.

## 8.2 Counting Animals – Capabilities and Challenges

Accurately counting and detecting animals was a primary concern and end-goal for many wildlife papers in the data corpus. This makes sense as wildlife surveys are a common task and critical tool for conservation efforts, land management, and wildlife research [21,93,119]. Traditional wildlife surveys are extremely costly, time-consuming, and even dangerous, depending on the habitats and species being studied [40]. Thus, automating any or all aspects of these surveys is highly desirable. Indeed, reviewed wildlife papers show continuous efforts throughout our fifteen-year review window to automate aspects of wildlife survey projects, such as using drones instead of manual data collection to automate data collection [40,63,70], and exploring automatic methods for animal detection, counting, and multi-species classification.

This review also found numerous papers in the PLF context focused on livestock counting [8,20,45,93,99,105,110,117,125,126]. However, unlike wildlife surveys, counting is only a first step towards the primary goal of animal health and welfare monitoring. Yet, there is still practical value in automating livestock inventories. Animal loss due to disease, predation, or rustlers is costly for farmers. Herds of sheep, cattle and other livestock who are raised on expansive ranges are often out-of-sight for days or weeks. Moreover, checking on them regularly is labour intensive, even dangerous, especially if livestock are in remote, rugged terrain. The ability to keep an accurate inventory of a herd could allow farmers to respond to a threat in a timely manner and limit losses.

Despite the plethora of reviewed papers that focused on animal counting – many with precision rates of nearly perfect – accurately and robustly doing so in many PLF contexts is still extremely difficult for vision-based techniques. Traditional computer vision approaches struggle with high variation in the background contexts of many animal habitats, including grass, sand, dirt, mud, rocks, and low-lying vegetation that can change with weather and with the seasons.

Variations in animal configuration, such as different poses, orientations, and social arrangements, all present challenges for accurately detecting individual animals in an image. Detecting smaller animals such as babies in a herd also posed a challenge for many reviewed techniques. Even in the best lighting conditions, these factors can lead to miscounting.

There was a clear trend in the literature towards adopting ML and DL methods to overcome these challenges to help improve animal counting in livestock and wildlife contexts. DL models show great promise for PLF applications. Yet, technical challenges remain for using these approaches for livestock counting. For new researchers in the field, it might be surprising to learn about the high accuracy of these models in recent research (e.g., 99.2% [9], 99.8% [14]). Yet, the reality is that these results are often based on "cleaned", fine-tuned data with noisy images removed for model training and testing. The accuracy of these models in real-world applications are likely much lower. Indeed, some reviewed papers included specific tests that utilized data collected from different locations than the training data, and achieved much lower detection performance scores on those datasets [107,110].

There are practical challenges that need to be overcome as well. Despite the fact that a 2018 review of the use of UAVs in livestock monitoring applications concluded that their use is likely "inevitable" given their potential utility and much larger use in the crop agriculture domain [7] these data capture platforms have real-world limitations. A key limiting factor for UAV applications on animal monitoring is their limited battery life, which is typically 30 minutes or less before recharging is needed [84].

## 8.3 Differentiating Animals – Capabilities and Challenges

Only a few reviewed papers focused on detecting and classifying different animal species. In a PLF context, multiple-species classification could be beneficial for farms that house different species or breeds of animals together. One paper explored the use of thermal imaging to differentiate between several large mammal species (elk, cattle, horses) with limited success [118]. More successful approaches (accuracies in the high nineties) used DL models for differentiating between different species of wildlife [35,106,115] or livestock [126]. Pretrained DL models have the ability to differentiate between various types of objects, as they are trained on hundreds of classes of objects. Thus, similar models to those discussed in section 12 would likely generalize to other contexts where the classification of different species or different breeds is desired. Employing transfer learning, where additional training is performed on the species or breeds of interest, may help improve model performance if target animals are different than those in the original training datasets. However, an ongoing challenge for developing highly effective DL models for livestock is the lack of sufficiently large datasets that would allow the development of general-purpose models, as discussed below.

## 8.4 Individual Animal Identification – Capabilities and Challenges

Addressing the ideal PLF goal of continuous 24/7 monitoring of individual livestock animals is predicated on the notion of being able to sense the state and behaviour of individual animals. Thus, vision-based monitoring must be able to identify individual animals. Yet very few reviewed papers focused on this task. Andrew et al. [3,4] used a DL model that combined the power of a CNN and analysis of information from prior video frames using the LRCN architecture to identify individual Holstein Friesian cattle, which have uniquely identifiable coat patterns. This method could be applied to livestock with differentiable features. Input data to these models must have sufficient detail to show discernable features. Thus, choosing an appropriate data capture method is important. For instance, capturing imagery at closer rather than farther distances, or potentially providing a side versus or complementary to a bird's eye view, may aid with the identification process.

In contrast, Schneider et al. [107] used similarity learning networks (Siamese and Triplet Loss networks), commonly used for facial recognition [58], to identify a variety of different animal species (whales, fruit flies, chimpanzees, tigers). Interestingly, these networks were able to identify individual fruit flies, which are indistinguishable with the human eye. This capability is promising for PLF applications because many livestock breeds have few observable differences in coat pattern or appearance. For instance, Angus cattle tend to be completely black with little to no discernable markings. Thus, investigation of similarity networks for individual livestock identification warrants further investigation.

Nevertheless, animal identification, especially in an outdoor setting, is an extremely challenging computer vision task, and may be best achieved using other methods. For instance, long-range passive ultra-high frequency (UHF) radio frequency identification (RFID) readers can be attached to a UAV to identify nearby cattle equipped with UHF-RFID ear tags [41]. Identification information could be stored with UAV imagery and used as meta data to assist image analysis.

## 9 DIRECTIONS FOR FUTURE RESEARCH

PLF researchers face many challenges to automate the whole livestock monitoring process in outdoor contexts, including but not limited to, selecting the right device to collect data, collecting data at the right time with proper background and foreground contrast and detail, and applying efficient algorithms to detect or count animals in real-time. The reviewed papers collectively offer substantial advance on these issues. However, more work is needed to establish best practices for different PLF contexts for vision-based livestock monitoring.

DL models show clear potential for livestock monitoring tasks. However, advances are needed in the areas of animal identification and detection of atypical animal behaviour or health states to reach the ideal PLF goal of individual animal health and welfare monitoring in these outdoor settings. Specific challenges our review uncovered are discussed below.

### 9.1 Encourage researchers to make datasets publicly available

DL models require extensive, carefully crafted data to develop robust, generalized models. The datasets used to train pre-trained models used by reviewed papers have hundreds of thousands (COCO) to millions (ImageNet) of labelled images. Many publicly available image datasets are available to foster research advances in general-purpose computer vision tasks and in specific application domain tasks. For example, annotated image datasets exist for facial recognition, object detection, medical imaging, pets, vehicles, and so on [1]. However, few such publicly accessible datasets exist for livestock species. Also, those datasets that do exist often contain idealized and "clean" data that does not sufficiently represent realistic scenes that would be captured in practice. For instance, the *AerialCattle* dataset [3] contains images of cattle in which only one cow is present, fully visible, and shown in profile with most of the background cropped out. These scenes are so different from imagery typically captured for PLF purposes, for instance, a bird's eye view of groups of cattle captured by a UAV, that models trained on the *AerialCattle* dataset would be almost useless in detecting cows in the UAV-captured dataset.

Consequently, many researchers whose work is integrated in the data corpus created their own datasets. This typically involved considerable time and effort to collect, clean, and label data to facilitate model training and testing. This represents weeks or months of work multiplied many times over for each research group. This necessarily slows potential progress in this field as time and money that could be spent on model development and refinement must be spent on dataset creation. Moreover, the use of custom datasets in these papers means that results of proposed models are not easily comparable. In contrast, if researchers make their datasets publicly available, others could use them to train and test new models more quickly, and also establish baselines upon which to compare different published models. Thus, we highly encourage researchers to make both their datasets and also their data annotations publicly available to help the whole field progress faster. Online file repositories such as *GitHub*[2] (subsidiary of Microsoft®) or ML communities, such as *Kaggle*[3] (subsidiary of Google®), can be used to publish open datasets for public use. We also encourage editors and reviewers in the PLF community to encourage authors of submitted papers to publish their datasets and funding bodies to support such dataset creation. Such efforts would increase the size and variety of livestock images available for testing new techniques, greatly reducing the time and effort needed to explore new models for livestock monitoring.

### 9.2 Consider motion detection and pose estimation

Some of the most unique livestock papers in the data corpus focused on motion and animal detection with contoured masks that matched the animal's profile. These approaches provide steps toward detecting abnormal behaviour or state of an animal. In the only motion detection paper found in this review [73], the goal was to explore the use of optical flow estimation to detect movement of a cow in a pasture environment (from the *AerialCattle* dataset). Although no further analysis was conducted in the paper, this was an important first step toward motion analysis, which could help detect potentially abnormal livestock behaviour. Unfortunately, motion data are extremely difficult to make as individual object pixels need to be mapped across video frames leading to solving a graph homomorphism problem, which is generally computationally challenging [34]. Traditional methods of capturing motion data or making motion datasets, for instance, attaching visible markers to moving objects or people or producing synthetic data, can be tedious, invasive, and expensive. Recent advances in the use of stereoscopic cameras to capture human motion [23], for instance, to conduct gait analysis [71], may be worth exploring for livestock applications.

Xu et al.'s [125,126] work on detecting cattle using the Mask R-CNN model was unique in its use of the Mask R-CNN DL approach [53] to produce contoured masks for detected objects referred to as image segmentation in the computer vision literature compared to more typical bounding boxes produced by other models. As Xu et al. suggest, this is an important first step towards activity analysis through, for instance, pose estimation. Indeed, PLF researchers have used Mask R-CNN together with the VGG-16 DL model to estimate pig poses in an indoor setting with moderate success (74% accuracy) to automatically detect pig behaviour [69]. The study used imagery with a top-down view of the pigs; however, the author suggests that choosing a camera angle where the pig's legs are visible may improve model performance.

### 9.3 Consider using post-processing techniques, ensemble models, and contextual models

A few reviewed studies employed DL models as an initial step in the animal detection process, whereby the model output was used to narrow the analysis space for further image processing. For instance, Barbedo et al. [8] used this approach to address observed limitations of CNNs to accurately detect small animals (calves) and distinguish between multiple clustered cows. This approach can help address known limitations of a given model.

Another approach to addressing individual model limitations is to employ the *ensemble* technique, whereby input data are sent to multiple models in parallel and a voting scheme is used to determine which model's results are used for the final prediction. Schneider et al.'s [106] used an ensemble model that included six DL models for multi-species classification. The ensemble model outperformed all individual models. This approach uses the respective advantages of different models, either due to training or model architecture, for different types of input data and may add robustness to livestock detection and related tasks.

Analysis of individual images or video frames can ignore valuable temporal context information that may improve detection outcomes. Many reviewed papers collected video data but then split the videos into individual video frames that were analyzed independently. In contrast, Andrew et al. [2] leveraged a model specifically designed to identify relevant context information between video frames, a long-term short-term memory (LSTM) model [57] and use that information to improve animal identification outcomes. These models are more complex to set up and use, but may help improve outcomes, especially if video data is already available.

---

[2] github.com
[3] kaggle.com

### 9.4 Consider using simulation approaches to create or augment datasets for training PLF DL models

The dearth of relevant datasets in many PLF contexts will be difficult to resolve using traditional time- and resource-intensive data capture methods. Building accurate DL models requires training datasets that capture animals under a wide variety of environmental conditions. Moreover, sufficiently representing rare but important events, such as cattle with specific health or welfare conditions, is important for accurate training. In similar sparce-data situations in other domains, simulated data, often referred to as synthetic data, are often used. For instance, edge cases for vision-based monitoring systems in autonomous vehicles have been created using virtual simulation environments and used to train models to detect such situations [18]. Using this approach, specific health or welfare issues could be simulated to help train PLF DL models.

Simulated approaches could also be used to create relevant datasets for motion-based monitoring. Liao et al.'s [73] livestock motion detection algorithms were tested on the MPI Sintel simulation-based dataset that is based on a 3D animated short film. Relevant PLF-relevant datasets are needed to progress motion-based monitoring in the field.

### 9.5 Consider using vision-based monitoring to support other livestock farming tasks

While PLF technologies tend to focus on individual animal outcomes, outdoor farming practices often focus on herd-level management, such as locating herds in open range environments [41] and habitat management [44]. Current capabilities for vision-based outdoor monitoring may already be useful for such tasks.

The ability to count animals also provides the means to *locate* them if, for instance, UAV platforms geotag images or stationary cameras can communicate information remotely. Animal detection systems can be combined with such location information to determine the location of a herd or an animal who has been separated from the herd. For livestock herds that cover large habitats, the ability to automatically locate and track herds over time and do regular herd inventories may be useful for farmers.

Maintaining pastures for livestock herds is an ongoing concern for many farmers who need to balance long-term sustainability of grazing areas with nutritional needs of their livestock. Tracking cattle's location in pasture areas over time could help farmers understand grazing behaviour. Moreover, image processing could be used to detect field conditions to help optimize pasture use. Prior work has explored the use of computer vision to estimate pasture quality [109] and to classify different types of pastures [46]. Combining such plant and soil-based detection with animal detection capabilities could help give a more complete understanding of pasture use.

García et al. [44] noted the lack of ML research in the area of environmental monitoring in livestock contexts, for instance, detection of dirty water that could potentially introduce illness to livestock. Droughts can also impact the availability of sufficient water supplies that put animals at risk. Vision-based remote sensing techniques have been used to monitor water quantity and quality in other contexts and may be worth exploring for PLF applications [129]

Raising safe and healthy livestock in a sustainable manner is a complex endeavour. Future research is warranted to develop comprehensive approaches to facilitate all aspects of animal and habitat monitoring.

### 10 CONCLUSION

This paper surveyed recent approaches for automated, vision-based outdoor animal monitoring in livestock and wildlife contexts. We uncovered many technical challenges inherent to the outdoor, unstructured nature of the task environment. We also found a clear trend toward the use of deep learning models to overcome these challenges. However, such models require large datasets to develop effectively. Such sizeable datasets are currently lacking in the livestock and wildlife fields. To mitigate this issue, researchers have begun exploring the use of large, pre-trained deep learning models, with promising results. Researchers using pre-trained "off-the-shelf" models or starting with these models and further refining them with local datasets using transfer learning approaches, saw significant improvements over traditional computer vision approaches and other machine learning approaches for animal monitoring related tasks.

Despite these advances, more work is still needed to move beyond the current animal monitoring capabilities, which largely consist of animal detection, counting, and to a limited extent, individual animal identification (when animals have clearly differentiable features). More cooperation is needed across the global PLF community to establish relevant open access benchmark datasets to reduce the need for costly data collection for each project undertaken in the field. Exploration of more advanced machine learning and deep learning techniques are also warranted, such as exploring ensemble models and temporal-based models that utilize contextual information details in video data. Notwithstanding the technical barriers uncovered in this review for vision-based outdoor monitoring of livestock, the current literature shows significant promise for herd-level monitoring in the near-term and individual animal monitoring in the future.

## 11 ACKNOWLEDGEMENTS

This work was funded in part by the Natural Sciences and Engineering Council of Canada (NSERC) and the University of Guelph. The authors would like to thank colleagues in the School of Computer Science at University of Guelph who provided valuable feedback on this work throughout the project.